\documentclass[sn-basic]{sn-jnl}

\usepackage{graphicx}%
\usepackage{multirow}%
\usepackage{amsmath,amssymb,amsfonts}%
\usepackage{amsthm}%
\usepackage{mathrsfs}%
\usepackage[title]{appendix}%
\usepackage{xcolor}%
\usepackage{textcomp}%
\usepackage{manyfoot}%
\usepackage{booktabs}%
\usepackage{algorithm}%
\usepackage{algpseudocode}
\usepackage{algpseudocode}%
\usepackage{listings}%

\numberwithin{equation}{section}
\theoremstyle{plain}

\newcommand{\iidsim}{\overset{\text{iid}}{\sim} }
\newcommand{\dd}{\mathop{}\! \mathrm{d}}

\theoremstyle{thmstyleone}%
\theoremstyle{thmstyletwo}%

\theoremstyle{thmstylethree}%

\begin{document}

\title[Simulation Based Bayesian Optimization]{Simulation Based Bayesian Optimization}


\author*[1]{\fnm{Roi} \sur{Naveiro}}\email{roi.naveiro@cunef.edu}

\author[2]{\fnm{Becky} \sur{Tang}}

\affil*[1]{\orgname{CUNEF Universidad}, \orgaddress{ \city{Madrid}, \postcode{28040}, \country{Spain}}}

\affil[2]{ \orgname{Middlebury College}, \orgaddress{\city{Middlebury}, \state{VT},  \postcode{05753},  \country{USA}}}


\abstract{Bayesian Optimization (BO) is a powerful method for optimizing black-box functions by combining prior knowledge with ongoing function evaluations. BO constructs a probabilistic surrogate model of the objective function given the covariates, which is in turn used to inform the selection of future evaluation points through an acquisition function. For smooth continuous search spaces, Gaussian Processes (GPs) are commonly used as the surrogate model as they offer analytical access to posterior predictive distributions, thus facilitating the computation and optimization of acquisition functions. However, in complex scenarios involving optimization over categorical or mixed covariate spaces, GPs may not be ideal. This paper introduces Simulation Based Bayesian Optimization (SBBO) as a novel approach to optimizing acquisition functions that only requires \emph{sampling-based} access to posterior predictive distributions. SBBO allows the use of surrogate probabilistic models tailored for combinatorial spaces with discrete variables. Any Bayesian model in which posterior inference is carried out through Markov chain Monte Carlo can be selected as the surrogate model in SBBO. We demonstrate empirically the effectiveness of SBBO using various choices of surrogate models in applications involving combinatorial optimization.}




\maketitle

\section{Introduction}

The problem of optimizing a potentially nonlinear function $f(\boldsymbol{x})$ over a certain set $\mathcal{S}$ has been extensively studied, see e.g. \cite{movckus1975bayesian}, \cite{jones1998efficient}. In many applications, the objective function is not explicitly known. This can make optimization challenging, as it is difficult to determine the optimal solution without accurate knowledge of the function. 
Moreover, $f(\boldsymbol{x})$ may be expensive to evaluate and a budget constraint often restricts evaluations of the objective function to a sparse selection of points, further complicating the optimization task. As an example, consider the process of drug discovery \citep{gallego2021ai}. During its early stages, the goal is to identify candidate molecules with a desired property, such as maximum binding affinity to a given biological target. However, measuring this binding affinity requires synthesizing and testing molecules \textit{in vitro}, limiting the number of leads that can be evaluated. Furthermore, expressing affinity as a mathematical function of the molecule structure is extremely difficult.

Bayesian Optimization (BO,  \cite{mockus1994application}) stands as an efficient approach for identifying the extrema of a black-box objective function. Its effectiveness originates from its capacity to integrate prior knowledge about the function with the evaluations previously conducted, enabling informed decisions on where to carry out additional evaluations.
BO involves the construction of a Bayesian surrogate model for $f(\boldsymbol{x})$. Subsequently, evaluation points are selected through the maximization of an acquisition function \citep{frazier2018tutorial}. 

Much of the BO literature focuses on continuous search spaces, where Gaussian Processes (GPs, \cite{gramacy2020surrogates}) are predominantly chosen as the Bayesian surrogate model. GPs are well-suited to optimization tasks in continuous search spaces as they rely on the smoothness defined by a kernel to model uncertainty.
Additionally, GPs provide an analytical posterior predictive distribution, allowing for efficient computation of acquisition functions like expected improvement or probability of improvement. However, GPs are less effective in more complex scenarios, such as those involving combinatorial (categorical) or mixed categorical-continuous search spaces, where the variables include both discrete and continuous components. This paper specifically addresses the challenge of adapting BO to perform effectively in such complex search spaces.


%
%
%
%
\paragraph{Related work} 

\noindent
In the last few years, there has been growing interest in performing Bayesian optimization over combinatorial domains. As highlighted by \cite{wang2023recent}, this task presents two major challenges: (i)~constructing surrogate models that capture the complexity of the combinatorial space and (ii)~efficiently searching that space to select the next point for evaluation. Most previous research has focused on the first challenge, often by retaining GP surrogates while designing specialized kernel functions adapted to discrete structures. In these approaches, the GP’s covariance function is modified to reflect similarity between discrete configurations rather than Euclidean distance. For example, \cite{garrido2020dealing} simply treat categorical (encoded via one-hot) and integer-valued variables as continuous and apply rounding before including them in the kernel. Other works rely on custom kernels based on Hamming distance or similar metrics (e.g., treating two configurations as more similar if they share more categorical components), thus aligning better with the discrete nature of the domain. Some methods employ string kernels (e.g., substring matching kernels) to model sequence data; for instance, \cite{moss2020boss} propose “BOSS”, a GP-based BO method that uses a subsequence string kernel to avoid naive one-hot encodings. Similarly, \cite{oh2019combinatorial} develop a GP with a diffusion kernel on a graph representing the combinatorial space. In their COMBO approach, each discrete value is treated as a vertex in a graph, and a diffusion-based covariance is defined over these vertices, enabling BO in high-dimensional binary and categorical inputs while preserving GP-based uncertainty. For an in-depth review of BO over discrete sequences, we refer the reader to \cite{gonzalez2025survey}.

\noindent
Beyond using fixed kernels, some methods combine GPs with learned embeddings or hybrid kernels to address more complex discrete domains. Instead of defining a kernel directly in the original combinatorial space, one can map discrete inputs into a learned continuous representation and then apply standard GP kernels. \cite{deshwal2021combining} follow this approach by using an autoencoder to embed sequences in a latent space, combining a latent-space kernel with a structured kernel in the original space. This hybrid kernel strategy enables the GP to capture both high-level features (through the latent mapping) and the finer discrete structure (through a dedicated combinatorial kernel). More recently, \cite{deshwal2023bayesian} propose a Hamming embedding via dictionaries, which preserves Hamming distances in a continuous space so that RBF or Matérn kernels can still reflect categorical similarity. Overall, a rich body of work has focused on retaining GPs while adapting their kernels to discrete search spaces.

\noindent
An alternative research direction replaces GPs with surrogates that more naturally accommodate high-dimensional or categorical inputs. One classical example is the use of random forests, as popularized by SMAC \citep{hutter2011sequential}, which discards the full GP posterior but scales more easily to many or purely categorical variables. Another category of GP alternatives uses parametric models, such as Bayesian linear regressions. For example, the proposal by \cite{baptista2018bayesian}, named BOCS (Bayesian Optimization of Combinatorial Structures), replaces the GP with a sparse Bayesian linear regression over polynomial expansions of binary inputs, placing a horseshoe prior on coefficients to identify a small subset of important interaction terms. This approach can be especially effective when the black-box function has mostly low-order interactions. Researchers have also investigated Bayesian neural network surrogates for high-dimensional BO, as in BOHAMIANN \citep{springenberg2016bayesian}, where a Bayesian neural network serves as the surrogate.

\noindent
In this paper, rather than designing a new surrogate model, we address the second challenge in BO for combinatorial spaces: efficiently searching a discrete domain for the next candidate once a suitable surrogate has been chosen. We introduce a novel simulation-based approach for optimizing acquisition functions in combinatorial domains, which we refer to as Simulation-Based Bayesian Optimization (SBBO). The key advantage of SBBO is its ability to optimize the acquisition function using only samples from the posterior predictive distribution, without requiring a closed-form expression. This makes BO applicable even when posterior inference is carried out through methods such as Markov Chain Monte Carlo (MCMC). By dispensing with closed-form assumptions, SBBO can leverage broader classes of surrogate probabilistic models that may be better aligned with the demands of combinatorial BO. This flexibility is particularly beneficial in high-dimensional discrete problems, where GPs are often criticized for poor scalability and the difficulty in defining meaningful kernel similarities as dimension increases.

The paper is structured as follows. Section \ref{sec:bo} provides an overview of Bayesian Optimization. Then, Section \ref{sec:sbbo} presents SBBO and summarises the surrogate probabilistic models that we use for the experiments presented in Section \ref{sec:exp}. Finally, Section \ref{sec:conclusions} gives some concluding remarks and lines of future research.

\section{Bayesian Optimization} \label{sec:bo}

Bayesian optimization is a class of methods centered on solving the optimization problem
\begin{equation}
\label{eq:opt}
    \boldsymbol{x}^{*} = \max_{\boldsymbol{x} \in \mathcal{S}} f(\boldsymbol{x}),
\end{equation}
In the BO setting, the \emph{objective function} $f(\boldsymbol{x})$ is assumed to be a black box, so no assumptions of convexity or linearity, for example, are made. Additionally, it is assumed that $f$ is expensive to evaluate in that the number of evaluations of $f$ that can be feasibly performed is small, perhaps due to computing time or financial cost. 
It is further assumed that querying the function at $x$ provides a noisy observation $y$ of $f(x)$.

BO uses a \emph{surrogate} probabilistic model of the objective function, usually obtained through Bayesian methods, to inform the next evaluation decision. It works iteratively by: 1) fitting the surrogate model, and 2) using an \emph{acquisition function} to obtain a new point $\boldsymbol{x}$ at which we evaluate or observe $f(\boldsymbol{x})$. The surrogate model is typically fitted using all available data $\mathcal{D}_{1:n} = \{(\boldsymbol{x}_{i}, y_{i}), i = 1, \ldots, n\}$. After selecting the new evaluation location $\boldsymbol{x}_{n+1}$, the corresponding $y_{n+1}$ is obtained and the surrogate model is updated to include the additional data point $(\boldsymbol{x}_{n+1}, y_{n+1})$. We now briefly review these two key components of Bayesian Optimization.

For the surrogate model, the following observational equation is assumed:
\begin{equation} \label{eq:obs_model}
    y  \equiv y(\boldsymbol{x}) = f(\boldsymbol{x}) + \epsilon, ~~~~~ \epsilon \iidsim \mathcal{N}(0, \sigma^2)
\end{equation}
where $f(\boldsymbol{x})$ could be parametric or non-parametric. Given a new location $\boldsymbol{x}$ and appropriate choice of prior, the posterior predictive distribution on $f(\boldsymbol{x})$ is taken as the surrogate model.
If a parametric model is chosen for $f$,  then $f(\boldsymbol{x})$ will depend on parameters $\boldsymbol{\beta}$ and is denoted as  $f(\boldsymbol{x}) \equiv f_{\boldsymbol{\beta}} (\boldsymbol{x})$. A prior $ \pi(\boldsymbol{\beta}) $ must be specified for $ \boldsymbol{\beta} $, which, when combined with the likelihood $ \pi(y \vert \boldsymbol{x}, \boldsymbol{\beta}, \sigma^2) $, yields the posterior $ \pi(\boldsymbol{\beta} \vert \mathcal{D}_{1:n}) $ after observing data $ \mathcal{D}_{1:n} $. The surrogate model is then the push-forward of this posterior through the mapping $ \boldsymbol{\beta} \mapsto f_{\boldsymbol{\beta}}(\boldsymbol{x}) $:
\begin{equation}\label{eq:ppd_param}
    \pi(f(\boldsymbol{x}) = z \vert \boldsymbol{x}, \mathcal{D}_{1:n}) \equiv  
    \int \delta \left(z - f_{\boldsymbol{\beta}} (\boldsymbol{x}) \right) \pi(\boldsymbol{\beta} \vert \mathcal{D}_{1:n}) \dd \boldsymbol{\beta}
\end{equation}
In most cases, the integral (\ref{eq:ppd_param}) lacks a closed-form solution. In these scenarios, surrogate model samples are typically generated by first drawing $ \boldsymbol{\beta} \sim \pi(\boldsymbol{\beta} \vert \mathcal{D}_{1:n}) $, often via MCMC, and then evaluating $ f_{\boldsymbol{\beta}}(\boldsymbol{x}) $.

If we instead choose a non-parametric model for $f$, then a prior $\pi(f)$ over $f$ needs to be specified. Coupled with the likelihood $\pi(y  | f, \sigma^2)$, we obtain the posterior distribution  $\pi(f | \mathcal{D}_{1:n} )$ over $f$. Then for a given new location $\boldsymbol{x}$, the evaluation of the posterior at $\boldsymbol{x}$, $\pi(f(\boldsymbol{x}) | \boldsymbol{x}, \mathcal{D}_{1:n})$, is taken as the surrogate model.

Regarding the acquisition function, there are several possible choices. We restrict our attention to acquisition functions that can be written as expectations of a utility function with respect to the posterior $\pi(f(\boldsymbol{x}) | \boldsymbol{x}, \mathcal{D}_{1:n} )$. Some of the most commonly used acquisition functions belong to this class, such as the \emph{expected improvement} and the \emph{probability of improvement}. Let $u(f(\boldsymbol{x}), \boldsymbol{x})$ denote the utility achieved by evaluating the function at a new point $\boldsymbol{x}$ and obtaining $f(\boldsymbol{x})$.
Then, the next point $\boldsymbol{x}_{n+1}$ is chosen as the one that maximizes the acquisition function, i.e. the expected utility:
\begin{equation} \label{eq:EI}
    \Psi(\boldsymbol{x}) \equiv \mathbb{E}_{f \vert \boldsymbol{x}, \mathcal{D}_{1:n} } \left [ u(f(\boldsymbol{x}), \boldsymbol{x}) \right].
\end{equation}

When using a GP prior for $f$, the posterior distribution is Gaussian, and for common utility functions, there is a closed-form expression for the expectation in Equation \eqref{eq:EI} that facilitates its optimization. The BO framework is flexible, but its computational efficiency and analytical results rely on this Gaussian process prior. However, in specific scenarios, such as when the search space includes categorical or mixed categorical-continuous variables, a GP prior may not be the most suitable choice. In these cases, alternative Bayesian models better suited to the problem may be available. However, these alternatives often lack an analytically-defined posterior distribution, complicating the selection of the next point 
$\boldsymbol{x}_{n+1}$ based on Equation \eqref{eq:EI}. Nonetheless, for most Bayesian models, we can sample from the posterior predictive distribution via MCMC methods. To address this challenge when employing a non-GP prior, we introduce a Bayesian optimization methodology that exclusively requires sampling access to the posterior. This approach enables the use of Bayesian models that are better suited to structured covariate spaces.

It is important to emphasize that choosing the next sampling location in \eqref{eq:EI} implicitly entails adopting a ``one-step" assumption: the selection is made under the presumption that no future evaluations will take place. When there is budget for more than one function evaluation, expected improvement and probability of improvement optimization unequivocally adopt a myopic approach. While several alternative strategies have been proposed for multi-step lookahead Bayesian Optimization \citep{wu2019practical, lam2016bayesian}, we focus on the one-step look-ahead case.

\section{Simulation Based Bayesian Optimization} \label{sec:sbbo}

We adapt ideas from the simulation-based optimal Bayesian design literature \citep{muller2005simulation} into the field of Bayesian Optimization, yielding our proposed method of Simulation-Based Bayesian Optimization (SBBO).

\subsection{Main Idea}

We aim to solve the optimization problem \eqref{eq:opt} with a limited number of function evaluations. Suppose we have access to data $\mathcal{D}_{1:n} = \{(\boldsymbol{x}_{i}, y_{i}), i = 1, \ldots, n\}$, possibly obtained from previous experiments, where $\boldsymbol{x}_{i}$ is a vector of covariates for observation $i$ and $y_{i}$ is derived from the observation model \eqref{eq:obs_model}. Given a budget of at most $M$ additional function evaluations, our task is to determine the next evaluation points $\boldsymbol{x}_{n+1}, \boldsymbol{x}_{n+2}, \ldots, \boldsymbol{x}_{n+M}$. As introduced in Section \ref{sec:bo}, these points are usually selected in a greedy manner. Following Bayesian decision-theoretic principles, the next evaluation should be performed at the location $\boldsymbol{x^*}$ that maximizes the expected utility \eqref{eq:EI}.

To perform Bayesian Optimization, we need to select a suitable Bayesian surrogate model and optimize \eqref{eq:EI}. As mentioned earlier, except in a few special cases, the posterior $\pi(f(\boldsymbol{x}) | \boldsymbol{x}, \mathcal{D}_{1:n} )$ is generally only accessible through sampling, typically using MCMC methods. When only sampling access to the posterior predictive distribution is available, a straightforward approach to optimizing \eqref{eq:EI} would be to construct a Monte Carlo (MC) approximation of the expected utility and optimize this approximation directly. However, this approach encounters significant difficulties. Because the expectation defining the acquisition function depends explicitly on the covariates $\boldsymbol{x}$, a separate MC approximation must be computed for every candidate solution explored during optimization, substantially increasing computational complexity. Moreover, as noted by \cite{muller2004optimal}, the expected utility surfaces tend to be particularly flat around their maxima, especially in high-dimensional scenarios. Consequently, MC approximation errors can obscure the true optimal regions, making them challenging to identify without an impractically large number of samples.

To address these limitations, we propose an alternative methodology drawn from the Bayesian decision analysis literature. 
Assuming we can generate\footnote{The specifics of the sampling process depend on the surrogate model employed (see Section \ref{sec:surrogate} for examples).} samples from $\pi(f(\boldsymbol{x}) | \boldsymbol{x}, \mathcal{D}_{1:n} )$, 
and that the utility function $u(f(\boldsymbol{x}), \boldsymbol{x})$ is positive and bounded\footnote{ Note that this positivity constraint is mild, as any bounded utility function can be shifted and scaled to become strictly positive without affecting the optimal decisions.};
we can borrow ideas from Bayesian decision analysis and reframe the optimization of \eqref{eq:EI} as a simulation problem. Following \cite{bielza1999decision}, we define an augmented distribution $g$ that treats $\boldsymbol{x}$ as a random variable:
\begin{eqnarray*}
    g(\boldsymbol{x}, f(\boldsymbol{x}) \vert \mathcal{D}_{1:n}) \propto  u(f(\boldsymbol{x}), \boldsymbol{x}) \pi( f(\boldsymbol{x}) | \boldsymbol{x}, \mathcal{D}_{1:n}).
\end{eqnarray*}

It is straightforward to see that the marginal of $g$ in $\boldsymbol{x}$ is proportional to the expected utility \eqref{eq:EI}. Thus, the mode of this marginal distribution coincides with the optimal next evaluation point. This suggests a simulation-based approach to find $\boldsymbol{x}_{n+1}$: we simulate $(\boldsymbol{x}, f(\boldsymbol{x})) \sim g (\boldsymbol{x}, f(\boldsymbol{x}) | \mathcal{D}_{1:n})$ using standard MCMC techniques and compute the mode of $g(\boldsymbol{x} | \mathcal{D}_{1:n})$ using the generated samples $\boldsymbol{x}$. 
However, this strategy is limited to low-dimensional $\boldsymbol{x}$. For high-dimensional covariate spaces, estimating the mode of $g(\boldsymbol{x} | \mathcal{D}_{1:n})$ from the simulation output becomes impracticable. \cite{muller2004optimal} propose an alternative augmented distribution $g_{H}$ defined as 
\begin{equation}\label{eq:augH}
    g_{H}(\boldsymbol{x}, f(\boldsymbol{x})_1, \ldots, f(\boldsymbol{x})_H | \mathcal{D}_{1:n}) \propto \prod_{h=1}^{H} u(f(\boldsymbol{x})_h, \boldsymbol{x}) \pi( f(\boldsymbol{x})_h | \boldsymbol{x}, \mathcal{D}_{1:n}),
\end{equation}
where $f(\boldsymbol{x})_1, \ldots, f(\boldsymbol{x})_H$ are $H$ copies of the random variable $f(\boldsymbol{x})$. Now, as the marginal of $g_{H}$ in $\boldsymbol{x}$ is
\begin{equation*}
    g_{H}(\boldsymbol{x} | \mathcal{D}_{1:n}) \propto \Psi(\boldsymbol{x})^{H},
\end{equation*}
the mode of this distribution also coincides with the optimal point at which we should make our next function evaluation. In particular, if $H$ is sufficiently large, the marginal distribution $g_{H}(\boldsymbol{x})$ tightly concentrates on the next $\boldsymbol{x}_{n+1}$, thus facilitating mode identification. 
Using this sharpened target is key to overcoming the challenges of direct optimization. When the acquisition function is relatively flat, naive Monte Carlo approximations are often ineffective, as stochastic errors can obscure the true peaks and cause the search to fail. In contrast, by sampling from a sharper version of the expected utility, our method concentrates probability density around the function's mode, providing a much more robust search.

\cite{muller2004optimal} propose an inhomogeneous MCMC sampling scheme to simulate from $g_H$. Their approach begins by generating samples $ (\boldsymbol{x}, f(\boldsymbol{x})) \sim g (\boldsymbol{x}, f(\boldsymbol{x}) | \mathcal{D}_{1:n})$ using the standard Metropolis-Hastings algorithm. In subsequent iterations, the target distribution is adjusted to $g_H$, with $H$ increasing incrementally. This inhomogeneous MCMC is designed so that the stationary distribution for a fixed $H$ is $g_H$.
We build upon this idea by adapting it to find the next evaluation point in the context of Bayesian optimization. Our proposed approach is detailed in Algorithm \ref{alg:sbbo}, where we utilize a Metropolis-Hastings sampling scheme. For clarity, we choose the state vector of the Markov chain to be $(\mathbf{x}, v)$ rather than $(\mathbf{x}, f(\mathbf{x})_1, \dots, f(\mathbf{x})_H)$, where $v = \frac{1}{H} \sum_{h=1}^H \log u(f(\mathbf{x})_h, \mathbf{x})$.

\begin{algorithm}[!htb]
\caption{ \hspace{0.5cm} Simulation Based Bayesian Optimization } \label{alg:sbbo}
\begin{algorithmic}
\State Obtain initial dataset $\mathcal{D}_{1:n} = \{ (\boldsymbol{x}_{i}, y_{i}), i = 1,\ldots, n \}$. Determine utility function $u_{\mathcal{D}}(\boldsymbol{x}, y)$.
\State Perform SBBO to find the next evaluation location.
\begin{enumerate}
    \item Choose initial $\boldsymbol{x}_{n+1}^{(0)}$. For $h = 1,\ldots, H$, simulate $f(\boldsymbol{x})^{(0)}_{h} \sim \pi(f(\boldsymbol{x}) | \boldsymbol{x}_{n+1}^{(0)}, \mathbf{D}_{1:n})$, evaluate utility $u^{(0)}_h = u(f(\boldsymbol{x}_{n+1}^{(0)})_h,\boldsymbol{x}_{n+1}^{(0)})$. 
    Define $v^{(0)} = \frac{1}{H}\sum_{h=1}^{H} \log u^{(0)}_h$.
    \item Generate candidate $\tilde{\boldsymbol{x}}_{n+1}$ from proposal distribution $q(\tilde{\boldsymbol{x}}_{n+1} | \boldsymbol{x}_{n+1}^{(0)})$.
    \item For $h = 1,\ldots, H$, simulate $\tilde{f}(\boldsymbol{x})_{h} \sim \pi(f(\boldsymbol{x}) | \tilde{\boldsymbol{x}}_{n+1}, \mathcal{D}_{1:n})$ and evaluate utility $\tilde{u}_{h} = u(\tilde{f}(\boldsymbol{x})_{h},\tilde{\boldsymbol{x}}_{n+1})$. Define $\tilde{v} = \frac{1}{H}\sum_{h=1}^{H} \log \tilde{u}_{h}$.
    \item Compute 
    \begin{align*}
    \alpha &= \min\left\{1, \exp\left(H \tilde{v} - H v^{(0)}\right) \cdot 
    \frac{q( \boldsymbol{x}^{(0)}_{n+1}| \tilde{\boldsymbol{x}}_{n+1})}{q(\tilde{\boldsymbol{x}}_{n+1} | \boldsymbol{x}^{(0)}_{n+1})}
    \right\}
    \end{align*}
    \item Set $$(\boldsymbol{x}_{n+1}^{(1)}, v^{(1)}) = \begin{cases}
    (\tilde{\boldsymbol{x}}_{n+1}, \tilde{v}) & w.p. \ \alpha \\
    (\boldsymbol{x}^{(0)}_{n+1}, v^{(0)}) & w.p. \ 1 - \alpha 
    \end{cases}$$
    \item Increase $H$ according to cooling schedule.
    \item Repeat (2) - (6) until convergence.
\end{enumerate}
\State Select $\boldsymbol{x}_{n+1}$ as the modal value of the simulated $\{\boldsymbol{x}_{n+1}^{(B)}, \boldsymbol{x}_{n+1}^{(B+1)}, \boldsymbol{x}_{n+1}^{(B+2)}, \ldots \}$, where the first $B$ simulated values are removed for burn-in.
\State Evaluate the true objective at $\boldsymbol{x}_{n+1}$ to obtain $y_{n+1}$. 
\State Update dataset $\mathcal{D}_{1:(n+1)} = \mathcal{D}_{1:n} \cup (\boldsymbol{x}_{n+1}, y_{n+1})$
\State Set $n = n+1$.
\end{algorithmic}
\end{algorithm}

If the utility is positive and bounded, and the state space of the inhomogeneous Markov chain generated by the sampler is finite, \cite{muller2004optimal} demonstrate that the chain is strongly ergodic when the update schedule for $H$ is logarithmic. Additionally, its stationary distribution is uniform over the set of $\boldsymbol{x}$ that maximize \eqref{eq:EI}. In practical situations where the state space is continuous or infinite, discretization of $v$ onto a finite grid becomes necessary.

In principle, as $H$ increases, the marginal distribution over $\boldsymbol{x}$ concentrates sharply around the global mode. Therefore, taking the last visited state of Algorithm~\ref{alg:sbbo} could theoretically yield the optimal evaluation point. However, in practice, to enhance numerical stability, we heuristically approximate the global mode by independently estimating the modal value for each covariate dimension from the simulated samples.

We have now transformed the optimization of \eqref{eq:EI} into a simulation problem. The objective is to solve this optimization without explicitly evaluating the posterior $\pi(f(\boldsymbol{x}) | \boldsymbol{x}, \mathcal{D}_{1:n} )$, relying only on samples from it. Therefore, we need to simulate from the augmented distribution \eqref{eq:augH} without requiring an analytical expression for the posterior.
This is where a key observation becomes crucial. Within the MCMC in Algorithm \ref{alg:sbbo}, we propose using $\pi(f(\boldsymbol{x}) | \boldsymbol{x}, \mathcal{D}_{1:n} )$ as the proposal distribution for $\tilde{f}(\boldsymbol{x})_h$ for $h = 1, \dots, H$, once a candidate $\tilde{\boldsymbol{x}}_{n+1}$ has been proposed. This approach ensures that the acceptance probability becomes independent of the posterior predictive density, allowing us to simulate from \eqref{eq:augH} without evaluating the posterior directly.
To see this more clearly: if the current state of the chain is $(\boldsymbol{x}, v)$, where $ v = \frac{1}{H} \sum_{h=1}^H \log u(f(\mathbf{x})_h, \mathbf{x}) $, the proposed new state is $(\tilde{\boldsymbol{x}}, \tilde{v})$, where $ \tilde{v} = \frac{1}{H} \sum_{h=1}^H \log u(\tilde{f}(\mathbf{x})_h, \tilde{\boldsymbol{x}}) $. With $q(\cdot | \boldsymbol{x})$ as the proposal distribution for $\boldsymbol{x}$, the acceptance ratio for each iteration of Algorithm \ref{alg:sbbo} is:

%
\begin{eqnarray*}
    \alpha &=& \min\left\{1,  
    \frac{ \prod_{h=1}^{H} u(\tilde{f}(\boldsymbol{x})_h, \tilde{\boldsymbol{x}}) \pi( \tilde{f}(\boldsymbol{x})_h | \tilde{\boldsymbol{x}}, \mathcal{D}_{1:n})   \pi( f(\boldsymbol{x})_{h} | \boldsymbol{x}, \mathcal{D}_{1:n})    }{\prod_{h=1}^{H} u(f(\boldsymbol{x})_{h}, \boldsymbol{x}) \pi( f(\boldsymbol{x})_{h} | \boldsymbol{x}, \mathcal{D}_{1:n})   \pi( \tilde{f}(\boldsymbol{x})_h | \tilde{\boldsymbol{x}}, \mathcal{D}_{1:n})  } \cdot \frac{q(\boldsymbol{x}| \tilde{\boldsymbol{x}})}{q(\tilde{\boldsymbol{x}} | \boldsymbol{x})}
    \right\} \\
    &=&  \min\left\{1,  
    \frac{ \prod_{h=1}^{H} u(\tilde{f}(\boldsymbol{x})_h, \tilde{\boldsymbol{x}})  }{\prod_{h=1}^{H} u(f(\boldsymbol{x})_{h}, \boldsymbol{x})   } \cdot \frac{q(\boldsymbol{x}| \tilde{\boldsymbol{x}})}{q(\tilde{\boldsymbol{x}} | \boldsymbol{x})}
    \right\} \\
    &=& 
    \min\left\{1, \exp(H \tilde{v} - H v )  \cdot
    \frac{q( \boldsymbol{x}| \tilde{\boldsymbol{x}})}{q(\tilde{\boldsymbol{x}} | \boldsymbol{x})}
    \right\}.
\end{eqnarray*}
Thus, the posterior predictive density for $\tilde{f}(\boldsymbol{x})$ cancels out with the proposal density, eliminating the need to evaluate it. This allows us to optimize \eqref{eq:EI} when just having sampling access to the posterior, thus achieving our goal.

\subsection{Gibbs Sampler}
\label{subsec:gibbs}
A possible variation of Algorithm \ref{alg:sbbo} entails using a Gibbs sampler to simulate from $g_H$ rather than a Metropolis-Hastings scheme for every $H$. Recall that $\boldsymbol{x}$ is a vector of $p$ covariates. The conditional distribution of the $s$-th element of $\boldsymbol{x}$ can be written as
\begin{equation*}
\begin{split}
    & g_H(\boldsymbol{x}_s \vert \boldsymbol{x}_{-s}, f(\boldsymbol{x})_1, \ldots, f(\boldsymbol{x})_H), \mathcal{D}_{1:n}) \propto  \\ & \quad  
     \exp \Bigg\lbrace  \sum_{h=1}^H \log \left[u( f(\boldsymbol{x})_h, \boldsymbol{x}_s \cup \boldsymbol{x}_{-s}) \right] + \log\left[ \pi(f(\boldsymbol{x})_h \vert \boldsymbol{x}_s \cup \boldsymbol{x}_{-s}, \mathcal{D}_{1:n} ) \right] \Bigg\rbrace
\end{split}
\end{equation*}
For the particular case in which $\boldsymbol{x}$ is a vector of $p$ categorical variables, this marginal is simply the \textit{softmax} distribution over $\sum_{h=1}^H \log \left[u( f(\boldsymbol{x})_h, \boldsymbol{x}_s \cup \boldsymbol{x}_{-s}) \right] + \log\left[ \pi(f(\boldsymbol{x})_h \vert \boldsymbol{x}_s \cup \boldsymbol{x}_{-s}, \mathcal{D}_{1:n} ) \right]$.  However, sampling from this distribution requires evaluating posterior predictive probabilities, which defeats the purpose of SBBO. The Gibbs sampler is thus beneficial only for scenarios in which the predictive distribution has a closed-form expression but the acquisition function does not. In such cases, the Gibbs sampler leverages the known predictive distribution to yield closed-form conditional marginals in the covariate space, significantly accelerating computation compared to a full Metropolis-Hastings scheme.

When closed-form predictive distributions are not available, we alternatively could create a Gibbs sampling scheme to sample from $g_H(\boldsymbol{x}_1, \boldsymbol{x}_2, \dots, \boldsymbol{x}_H | \mathcal{D}_{1:n})$ where $(f(\boldsymbol{x})_1, \ldots, f(\boldsymbol{x})_H)$ has been integrated out. Sampling from $g_H(\boldsymbol{x}_s \vert \boldsymbol{x}_{-s}, , \mathcal{D}_{1:n})$ can be done using Metropolis steps. Assuming that the current state of the chain is $\boldsymbol{x}$, $f(\boldsymbol{x})_1, \dots,f(\boldsymbol{x})_H $ and $v = \frac{1}{H} \sum_{h=1}^H \log u(f(\boldsymbol{x})_h, \boldsymbol{x} )$, we proceed as follows:
\begin{enumerate}
    \item Propose a candidate $\tilde{\boldsymbol{x}}_s$ using a probing distribution $q_s(\tilde{\boldsymbol{x}} \vert \boldsymbol{x})$ that only changes the $s$-th element of $\boldsymbol{x}$.
    \item For $h = 1, \ldots, H$, simulate $\tilde{f}(\boldsymbol{x})_h \sim \pi(f(\boldsymbol{x}) \vert \tilde{\boldsymbol{x}}_s \cup \boldsymbol{x}_{-s}, \mathcal{D}_{1:n})$. 
    \item Compute $\tilde{v} = \frac{1}{H} \sum_{h=1}^H \log u(f(\tilde{\boldsymbol{x}}_s \cup \boldsymbol{x}_{-s})_h, \tilde{\boldsymbol{x}}_s \cup \boldsymbol{x}_{-s} )$
    \item Evaluate acceptance probability
    \begin{align*}
    \alpha &= \min\left\{1, \exp(H \tilde{v} - H v ) 
    \frac{q( \boldsymbol{x} |\tilde{\boldsymbol{x}}_s \cup \boldsymbol{x}_{-s})}{q(\tilde{\boldsymbol{x}}_s \cup \boldsymbol{x}_{-s} | \boldsymbol{x} )}
    \right\}
    \end{align*}
\end{enumerate}
This Gibbs sampling scheme would be embedded in the inhomogenous Markov chain defined in Algorithm \ref{alg:sbbo}.

SBBO allows us to naturally accomodate the scenario of mixed categorical-continuous covariate spaces. We simply adapt the proposal distribution of each particular element. For instance, if $\boldsymbol{x}_q$ is continuous we could use a Gaussian proposal distribution, while if it is categorical we could use a discrete uniform proposal over all possible choices for the covariate. 

Another possible variation aside from the Gibbs sampler is to use Sequential Monte Carlo (SMC) methods to approximate both the posterior distribution and the expected utility. For instance, \cite{oliveira2022bayesian} proposed employing SMC to sequentially approximate the posterior over the parameters of a surrogate parametric model in Bayesian optimization, subsequently computing expected utilities as weighted sums over the resulting particles. In our context, however, an even more suitable approach could be the interacting particle system proposed by \cite{amzal2006bayesian}, which samples from progressively annealed versions of the augmented distribution \eqref{eq:augH}, gradually concentrating particles around its mode. Adapting this sampling strategy to Bayesian optimization represents an intriguing direction for extending the SBBO framework.

\subsection{Probabilistic Surrogate Models} \label{sec:surrogate}

We can now optimize \eqref{eq:EI} for any probabilistic surrogate model, as long as we have access to posterior samples of $f$. This flexibility allows us to consider a wide range of surrogate models. In particular, we have tested the following models, which may be well-suited for cases where $\mathcal{S}$ is the space of $p$-dimensional categorical covariates.

\paragraph{Tanimoto Gaussian Process} This surrogate model, introduced by \cite{moss2020gaussian}, is particularly useful when $x \in \lbrace 0, 1 \rbrace^p$. The uncertainty on $f(x)$ is modeled through a Gaussian Process with the following kernel function:
\begin{equation*}
k(x, x') = \frac{ \phi \cdot x^\top \cdot x' }{\Vert x \Vert^2 + \Vert x' \Vert^2 - x \cdot x'}.
\end{equation*}
The numerator is proportional to the number of dimensions where both $x$ and $x'$ are 1, and the denominator is the total number of dimensions where either $x$ or $x'$ or both is 1. The parameter $\phi$ is usually set to the value maximizing the marginal likelihood, thus following a empirical Bayes approach.
For categorical covariates with more than two possible values, we would need to first convert the covariates into appropriate binary variables (e.g., via one-hot encoding) before computing the previous kernel. Under this model, the posterior predictive distribution is Gaussian with mean and variance analytically available as in any GP. 

\paragraph{Sparse Bayesian linear regression}

The response variable is modeled as a linear function of the covariates, including all possible pairwise interactions. A heavy-tailed horseshoe prior is used to induce sparsity.
\begin{eqnarray}
&& y = f_{\boldsymbol{\alpha}}(\boldsymbol{x}) + \epsilon = \alpha_0 + \sum_j \alpha_j x_j + \sum_{i,j>i} \alpha_{ij} x_i x_j + \epsilon \label{eq:bocs}\\
&& \epsilon \sim \mathcal{N}(0, \sigma^2) \nonumber \\ 
&& \alpha_k \vert \beta_k, \tau, \sigma^2 \sim \mathcal{N}(0, \beta_k^2 \tau^2 \sigma^2) \nonumber\\
&& \beta_k, \tau \sim \text{Cauchy}_{[0,\infty)}(1) \nonumber\\
&& p(\sigma^2) \propto \sigma^{-2} \nonumber
\end{eqnarray}
This model was first proposed by \cite{baptista2018bayesian} in the context of Bayesian Optimization, where posterior inference is performed using Gibbs sampling. This method allows us to sample from the posterior of $\alpha_k, \beta_k, \tau$, and $\sigma^2$, facilitating easy draws from the posterior of $f_{\alpha}(\boldsymbol{x})$ that could be directly utilized in Algorithm \ref{alg:sbbo}.
In this scenario, access to analytical posterior predictive distributions is unavailable, making the computation of common acquisition functions challenging. To address this, \cite{baptista2018bayesian} propose a tailored acquisition function: a single posterior sample of $\alpha$ is produced, and the acquisition function is defined as $f_{\alpha}(\boldsymbol{x}) - \lambda \Vert \boldsymbol{x} \Vert_1$, where a penalty term controlled by $\lambda$ is included. This acquisition function now has a closed-form, appearing as a quadratic form. Its optimization can be reformulated as a semidefinite program that can be efficiently approximated in polynomial time to a desired precision.
With SBBO, there is no need for tailored acquisition functions as we just require having sampling access to the posterior predictive distribution. This enables using model \eqref{eq:bocs} with the common acquisition functions in the BO literature. It is true though that the theoretical results presented in Section 3.1 require samples from the posterior predictive distribution to be independent and identically distributed, a condition typically not satisfied by MCMC-generated samples. Nevertheless, we find that thinning MCMC samples to reduce autocorrelation provides good empirical performance, as demonstrated in the experiments section.


\paragraph{NGBoost}
Natural Gradient Boosting (NGBoost, \cite{duan2020ngboost}) is a non-Bayesian algorithm for generic probabilistic prediction via gradient boosting. In NGBoost, the conditional distribution of the output given covariates is modeled through a parametric distribution $f(\boldsymbol{x}) \vert \boldsymbol{x} \sim P_\theta (\boldsymbol{x})$. The parameters $\theta$ are obtained through an additive combination of $M$ base learners $b^{(m)}$ and an initial $\theta^{(0)}$,
$$
\theta = \theta^{(0)} - \eta \sum_{m=1}^M \rho^{(m)}\cdot b^{(m)} (x)
$$
where $\rho^{(m)}$ are scaling factors and $\eta$ is a common learning rate.
The base learners are trained to minimize a proper scoring rule using a refinement of gradient boosting. For this training process, NGBoost employs the natural gradient, a variant of gradient descent that takes into account the underlying geometry of probability distributions leading to faster convergence and improved stability.

A notable strength of NGBoost lies in its adaptability, as it can be paired with various base learners. Our experiments focus on two such learners: shallow decision trees (referred to as NGBoost-dec) and linear regressions with lasso regularization (named NGBoost-linCV). For both setups, we use the log-score as the scoring rule, with the distribution $P_\theta$ set to a normal distribution in all experiments.

\paragraph{Bayesian Neural Network}
Bayesian Neural Networks (BNNs) are probabilistic deep learning models that replace deterministic weight values of a neural network with weight distributions. This allows the model to capture the existing epistemic uncertainty in the predictions due to limited training data. In our experiments, we employ a simple three hidden layer fully-connected BNN whose inputs are the one-hot encoded covariates. We train this model using variational inference, combined with the reparametrization estimator technique \citep{kingma2013auto}.

\section{Experiments}\label{sec:exp}

We conduct experiments in three different combinatorial optimization problems\footnote{All code to reproduce the experiments in this paper has been open sourced and is available at \url{https://github.com/roinaveiro/sbbo}.}: a binary quadratic program with 10 variables, a contamination control in a food supply chain with 25 stages, and RNA sequence optimization with a sequence length of 30. We evaluate the performance of SBBO with the four probabilistic surrogate models from Section \ref{sec:surrogate}: Tanimoto Gaussian Process Regression (sbbo-GPr), Sparse Bayesian linear regression with interactions (sbbo-BLr), natural gradient boosting with a shallow decision tree model as base learner (sbbo-NGBdec), natural gradient boosting with an sparse linear regression as base learner (sbbo-NGBlinCV) and Bayesian Neural Network (sbbo-BNN).
We compare the performance of these SBBO approaches with three benchmarks: simulated annealing (SA), a widely used algorithm in combinatorial optimization problems \citep{spears1993simulated}, random local search (RS, \cite{bergstra2012random}) and COMBO, a state-of-the-art Bayesian optimization method specifically designed for combinatorial domains introduced by \cite{oh2019combinatorial}.

The experimental setup was as follows. We start with a random initial dataset containing five pairs of covariates and responses. For each probabilistic surrogate model and fixed number of iterations $t$, we follow these steps:
\begin{enumerate}
    \item Use the surrogate model to determine the next evaluation point, employing the expected improvement as the acquisition function. Thus, the utility function is $u(\boldsymbol{x}) = \max \left[ f(\boldsymbol{x}) - f^*(\boldsymbol{x}), 0 \right]$, where $f^*(\boldsymbol{x})$ represents the current estimate of the optimal true objective. As SBBO algorithm, we employ a Metropolis-Hastings sampling scheme, as in Algorithm \ref{alg:sbbo}. The proposal distribution randomly selects a covariate and assigns it a new value, uniformly sampled from its possible range.
    The cooling schedule for $H$ is updated at each iteration  of Algorithm 1. For our experiments, we employ a linear schedule where $H$ starts at 1 and is increased by 250 at each subsequent iteration. This more aggressive schedule was found to perform well in practice, though other schedules could be explored.
    \item Once the next evaluation point is determined, query the black box function to obtain a noisy observation of the true objective at that location.
    \item Append the new pair of data to the dataset and repeat the process.
\end{enumerate}
For each algorithm, we report the best function value obtained after $t$ evaluations of the true objective, averaged over 10 runs with the corresponding one standard error interval. The value of $t$ is chosen based on the specific problem setting. Note, however, that the COMBO algorithm differs from the other methods in that its acquisition function optimization is deterministic. Thus, COMBO results do not include uncertainty quantification.

\subsection{Binary Quadratic Problem}

\begin{table}[htb]
\centering
\caption{Average objective function values and margin of error for each algorithm after 120 iterations for the binary quadratic problem, 500 iterations for the contamination control problem and 300 iterations for the RNA problem. Bold values denote  best value for each problem.}
 \label{tab:full}
\begin{tabular}{lccc}
\toprule
\textbf{Algorithm} & \textbf{BQP (120 Iter.)} & \textbf{CON (500 Iter.)} & \textbf{RNA (300 Iter.)} \\
\midrule
RS & $ 9.86 \pm 0.27 $ & $ 21.74 \pm 0.04 $ & $ -13.74 \pm 0.63 $ \\
SA & $ 10.75 \pm 0.33 $ & $ 21.44 \pm 0.04 $ & $ -9.21 \pm 0.61 $ \\
COMBO & $\boldsymbol{11.24 \pm 0.00 }$ & $\boldsymbol{ 21.19 \pm 0.00 }$ & $ -17.00 \pm 0.00 $ \\
sbbo-BLr & $ \boldsymbol{11.24 \pm 0.00} $ & $ 21.26 \pm 0.01 $ & $ \boldsymbol{-22.65 \pm 0.59} $ \\
sbbo-GPr & $ 10.89 \pm 0.14 $ & $\boldsymbol{ 21.19 \pm 0.00 }$ & $ -15.05 \pm 0.92 $ \\
sbbo-BNN & $ 10.42 \pm 0.30 $ & $ 21.50 \pm 0.02 $ & $ -16.08 \pm 0.56 $ \\
sbbo-NGBlinCV & $ 10.47 \pm 0.27 $ & $ 21.37 \pm 0.03 $ & $ -14.04 \pm 0.62 $ \\
sbbo-NGBdec & $ 10.75 \pm 0.25 $ & $ 21.38 \pm 0.02 $ & $ -13.82 \pm 0.59 $ \\
\bottomrule
\end{tabular}
\end{table}

The aim here is to maximize a quadratic function incorporating $l_1$ regularization. The objective function is expressed as $\boldsymbol{x}^\top Q \boldsymbol{x} - \lambda \Vert \boldsymbol{x} \Vert_1$, where $\boldsymbol{x}$ is a binary vector of dimension $d=10$. The matrix $Q$ is a random $d \times d$ matrix with entries generated as follows: each element is simulated as independent standard Gaussian with the resulting matrix then element-wise multiplied by another $d \times d$ matrix $K$, whose elements are $K_{ij} = \exp\left[ -(i-j)^2 / L_c^2\right]$, with the correlation length $L_c^2$ governing how rapidly the values of $K$ decay away from the diagonal. A larger value of $L_c^2$ results in a denser matrix $Q$, complicating the optimization task. In our experiments, we set $\lambda$ to 0 and fixed $L_c^2$ at 10.
\begin{figure}[htb]
    \centering 
    \includegraphics[width=0.9\textwidth]{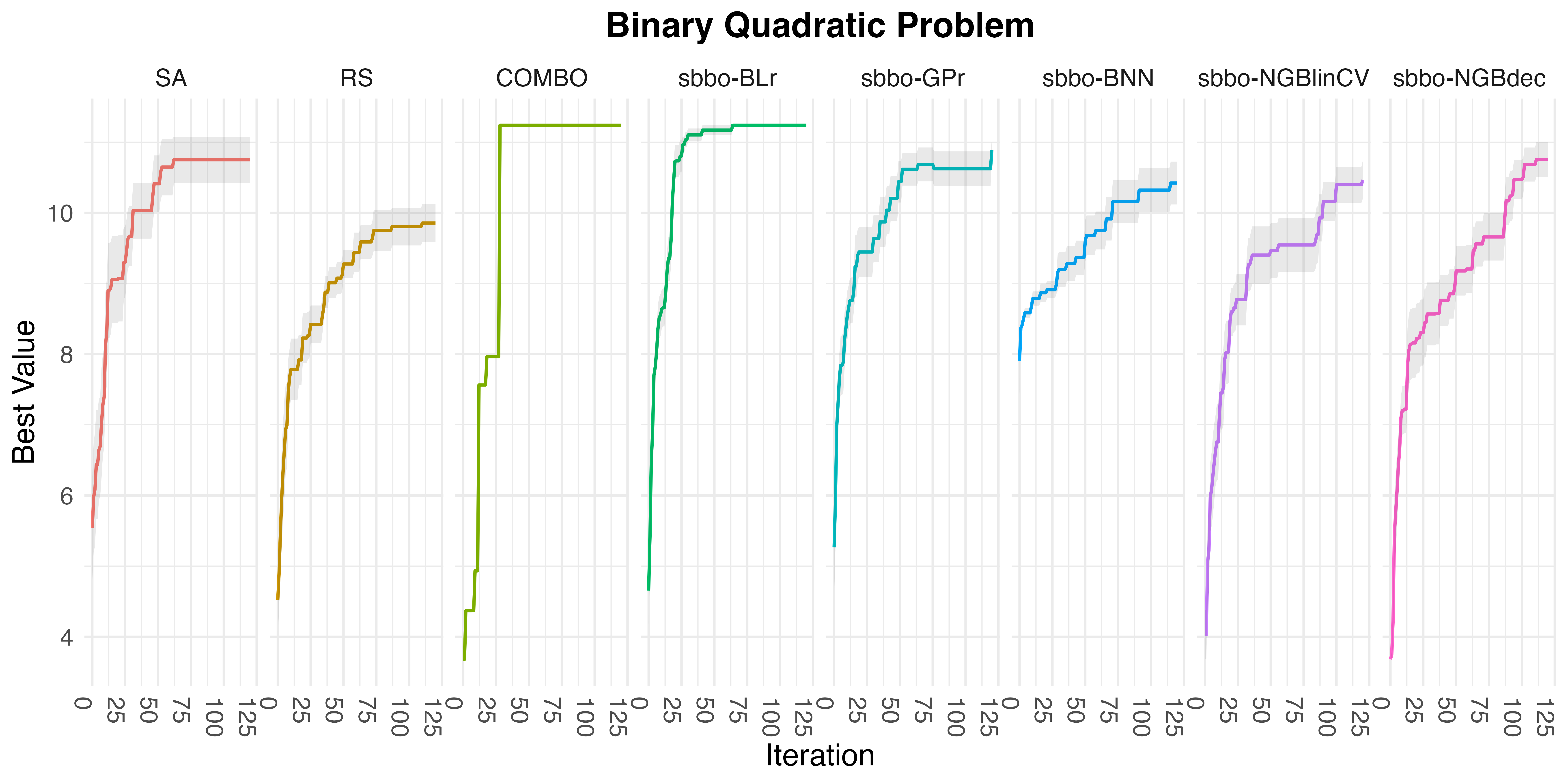}
    \caption{Binary quadratic problem: average $\pm$ one standard error of objective function values by number of function evaluations. Large objective values indicate better performance.}
    \label{fig:bqp}
\end{figure}
The performance comparison of the different algorithms over 120 iterations is illustrated in Figure \ref{fig:bqp} and detailed in the second column of Table \ref{tab:full}. As can be seen, SBBO outperforms RS and SA for most surrogate models.
Notably, COMBO and SBBO with sparse Bayesian linear regression achieves the highest objective value.
An essential aspect is the assessment of performance with a limited number of objective function evaluations, as evaluating the true objective is the common bottleneck in most BO applications. Remarkably, COMBO, SBBO-BLr, and SA demonstrate good performance for a low number of evaluations ($<$25) (Fig. \ref{fig:bqp}). However, it is worth noting that SA appears to get stuck in local optima around the 65th evaluation. 

In this experiment, with a search space cardinality of 1024, it is feasible to compute the global optimum (11.24) via complete enumeration. Table \ref{tab:bqp} displays the deviation between the optimal solution discovered within 120 function evaluations for each algorithm and the global optimum. Notably, sbbo-BLr and COMBO consistently identifies the optimal solution within 120 $\ll$ 1024 iterations across all 10 repetitions of the experiment.

\begin{table}
\caption{Binary quadratic problem: average distance to global optimum after 120 iterations with margin of error. Bold indicates best average performance.}
\label{tab:bqp}
\centering
\begin{tabular}[t]{lc}
\toprule
\textbf{Algorithm} & \textbf{Distance to global optimum}\\
\midrule
RS & $ 1.38 \pm 0.27 $\\
SA & $ 0.49 \pm 0.33 $\\
COMBO & $\boldsymbol{ 0.00 \pm 0.00 } $\\
sbbo-BLr & $ \boldsymbol{0.00 \pm 0.00} $\\
sbbo-GPr & $ 0.35 \pm 0.14 $\\
sbbo-BNN & $ 0.82 \pm 0.30 $\\
sbbo-NGBlinCV & $ 0.77 \pm 0.27 $\\
sbbo-NGBdec & $ 0.49 \pm 0.25 $\\
\bottomrule
\end{tabular}
\end{table}

\subsection{Contamination Control Problem} \label{sec:cont_problem}

The contamination control problem was proposed by \cite{hu2010contamination}. Consider a food supply with $d=25$ stages that may be contaminated with pathogenic microorganisms. $Z_i$ denotes the fraction of food contaminated at the $i$-th stage of the supply, for $1 \leq i \leq d$. $Z$ evolves according to the following random process. At stage $i$, a prevention effort (with fixed cost $c_i$) can be made that decreases contamination a random rate $\Gamma_i \in (0,1)$. 
If no prevention is taken, contamination spreads at a random rate $\Lambda_i \in (0,1)$. This results in the recursive equation
\begin{equation*}
Z_i = \Lambda_i (1-x_i)(1 - Z_{i-1}) + (1 - \Gamma_i x_i) Z_{i-1}
\end{equation*}
where $x_i \in \lbrace 0,1 \rbrace$ is the decision variable associated with the prevention effort at stage $i$, and $x_{i} = 1$ denotes the the prevention effort was taken. The goal is to decide, at each stage $i$, whether to implement a prevention effort. The problem is constrained in that the cost should be minimized while ensuring that at each stage, the fraction of contaminated food does not exceed an upper limit $U_i$ with probability at least $1-\epsilon$. The random variables $\Gamma_i$ and $\Lambda_i$ as well as the initial fraction of contaminated food $Z_1$ follow beta distributions. We set $U_i = 0.1$ for all $i$ and $\epsilon=0.05$, thus adopting the same framework as in \cite{baptista2018bayesian}.

We consider the Lagrangian relaxation of the problem given by
\begin{equation*}
\arg\min_x \sum_{i=1}^d \left[ c_i x_i + \frac{\rho}{T} \sum_{k=1}^T \left(1_{\lbrace Z_{ik} > U_i \rbrace} - (1- \epsilon)\right)\right] + \lambda \Vert x \Vert_1
\end{equation*}
where each violation is penalized by $\rho=1$. We set $T=100$. An $l_1$ regularization term with parameter $\lambda = 0.0001$ is added to encourage the prevention efforts to occur at a small number of stages.
\begin{figure}[htb]
    \centering
    \includegraphics[width=\textwidth]{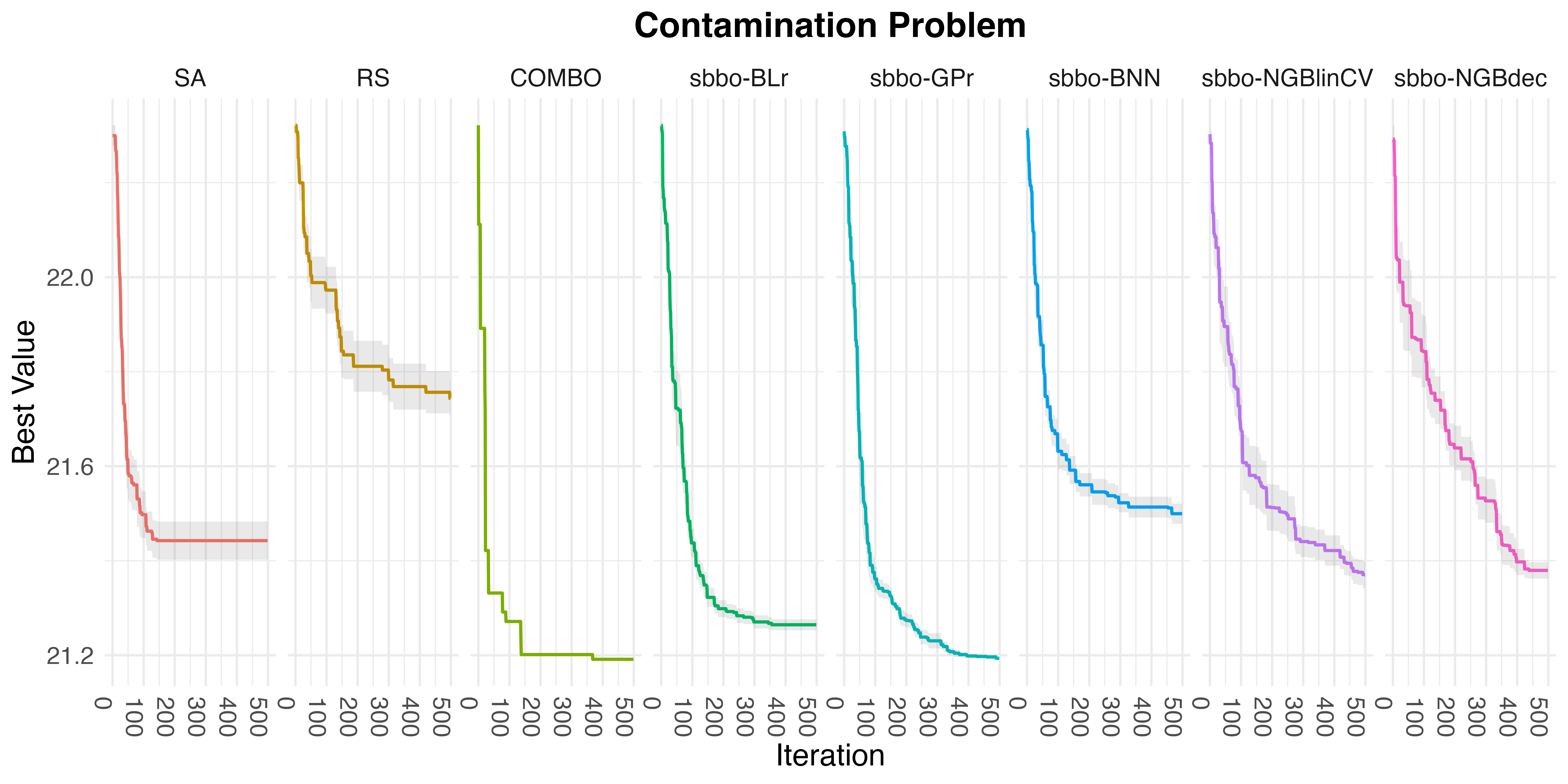}
    \caption{Contamination control problem: average $\pm$ one standard error of objective function values by number of function evaluations. Small objective values indicate better performance.}
    \label{fig:con}
\end{figure}
The performance comparison of the different algorithms over 500 iterations is illustrated in Figure \ref{fig:con} and detailed in the third column of Table \ref{tab:full}. 
For a small number of function evaluations, simulated annealing shows strong initial performance but quickly becomes trapped in a local optimum. COMBO, on the other hand, demonstrates consistent performance across both low and high numbers of function evaluations, achieving the most accurate estimation of the global minimum within 500 evaluations. SBBO-GPr closely follows COMBO and also achieves the best function value after 500 iterations, matching COMBO's performance at convergence.

Importantly, above $\sim 350$ function evaluations, all SBBO-based methods except for SBBO-BNN outperform SA and RS.

\subsection{RNA design}

An RNA sequence is represented as a string $A = a_1 \cdots a_p$ of length $p$, where each $a_i \in \lbrace A, U, G, C \rbrace$ represents a nucleotide. The secondary structure of the RNA is defined by the arrangement of base pairs that minimizes its free energy, referred to as the Minimum Free Energy (MFE). The objective of RNA sequence optimization is to find the sequence that folds into the structure with the lowest MFE.

Several RNA folding algorithms, such as the one introduced by \cite{zuker1981optimal}, use dynamic programming and thermodynamic models to estimate the MFE of a sequence. However, these algorithms have a computational complexity of $\mathcal{O}(p^3)$, which makes them inefficient for evaluating a large number of sequences. The RNA sequence optimization problem becomes even more challenging as the number of possible sequences grows exponentially with $4^p$, making exhaustive search for the global minimum MFE impractical.

To overcome these challenges, we use BO to efficiently search for the optimal RNA sequence, reducing the number of MFE evaluations required. We use the \texttt{RNAfold} package \cite{lorenz2011viennarna} to compute the MFE for a given sequence. The results presented focus on sequences of length $p = 30$.




\begin{figure}[htb]
    \centering 
    \includegraphics[width=0.9\textwidth]{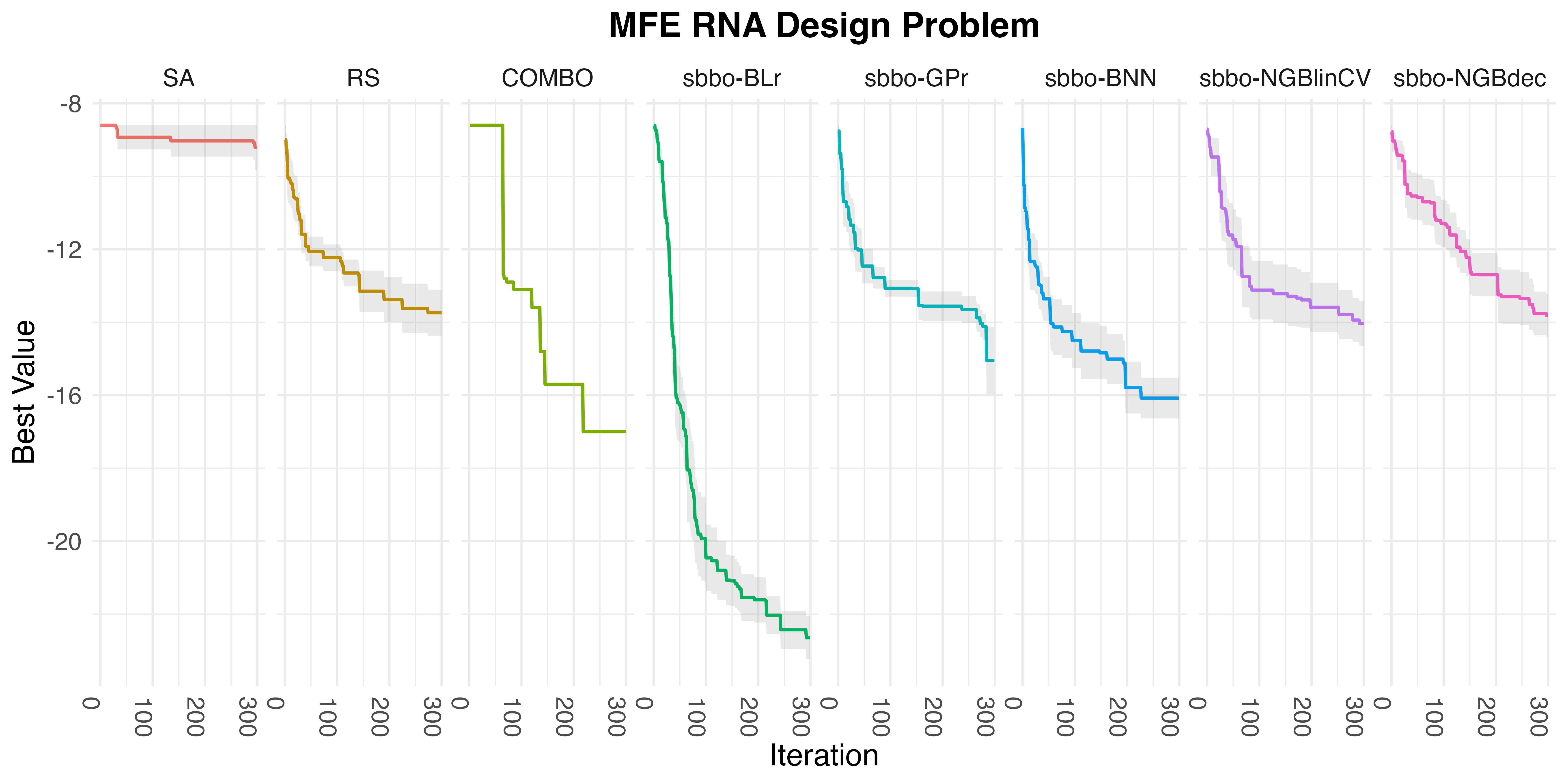}
    \caption{RNA problem: average $\pm$ one standard error of objective function values by number of function evaluations. Small objective values of MFE indicate better performance.}
    \label{fig:rna}
\end{figure}

Figure \ref{fig:rna} and the last column of Table \ref{tab:full} showcase the performance comparison of various algorithms over 300 iterations. Notably, Simulated Annealing significantly under-performs, even when compared to Random Local Search, possibly due to the extensive search space (cardinality of $2^{60}$). Remarkably, SBBO-BLr surpasses all other methods by a significant margin. While COMBO and SBBO using a Bayesian Neural Network surrogate both perform well for a small number of function evaluations and achieve the second- and third-best results after 300 iterations respectively, their performances remain clearly behind SBBO-BLr. These findings highlight the potential benefits of exploring alternative Bayesian surrogate models beyond Gaussian processes, particularly in complex, high-dimensional, combinatorial optimization tasks.

\subsection{General Comments}

While most Bayesian Optimization literature have utilized Gaussian Processes as surrogate probabilistic models, our empirical study suggests that, for combinatorial optimization problems, alternative models may lead to superior performance. Notably, in two out of three experiments, Sparse Bayesian linear regression with pairwise interactions demonstrated clear superiority over GPs and maintained competitiveness in the remaining experiment. This underscores the significance of exploring a broader range of Bayesian models within the realm of Bayesian Optimization.

Given that many Bayesian models lack closed-form expressions for posterior predictive distributions, our study emphasizes the relevance of developing algorithms to optimize acquisition functions in Bayesian Optimization when only sampling access to posteriors is available. This is particularly important as inference in most Bayesian models relies on simulation. SBBO aligns precisely with this objective.

\section{Conclusion and Future Work}\label{sec:conclusions}

We have proposed SBBO, a simulation-based framework for optimizing acquisition functions that only requires sampling access to a surrogate model's posterior predictive distribution. The key advantage of SBBO is its generality, as the framework can be used with any Bayesian surrogate model for which posterior predictive sampling is feasible. Based on our analysis and experience, SBBO is most appropriate for: (i) high-dimensional problems; (ii) scenarios where log-predictive densities are intractable, precluding alternative methods; and (iii) cases where the acquisition function is suspected to be flat near its maxima, making direct optimization challenging. We have demonstrated its empirical effectiveness across several surrogate models in three combinatorial optimization problems.

Several future lines of research are possible. From a methodological perspective, we have only considered the greedy one-step look-ahead BO. SBBO could be extended to multi-step look-ahead BO. However, this extension is not direct. From a computational perspective we would suffer from the standard inefficiencies of dynamic programming. Some ideas in Bayesian sequential design might be relevant \cite{wathen2006implementation, brockwell2003gridding, drovandi2014sequential}.
In terms of sampling strategies, we have proposed Metropolis-Hastings and Gibbs samplers to sample from the augmented distribution. Other options could be adopted to improve exploration of high-dimensional multimodal surfaces, such as particle methods \cite{amzal2006bayesian}. In addition, when dealing with continuous search spaces, MCMC samplers that leverage gradient information, such as Hamiltonian Monte Carlo or the Metropolis-adjusted Langevin algorithm could be used.

On the applications side, given that Bayesian linear regression with pairwise interactions seems to work well, a model that captures interactions between three or more covariates is of interest. However, explicitly accounting for beyond-pairwise interactions becomes unfeasible when the dimensionality of the search space increases. As an alternative, some non-parametric approaches such as Bayesian additive regression tress \citep{chipman2010bart} that capture relevant interactions between more than two covariates could be utilized.
Finally, SBBO could be used with likelihood-free surrogate models, from which posterior predictive samples could be obtained via Approximate Bayesian Computation.

\section*{Acknowledgements}

This project is partially supported by the Spanish state investigation agency under  the \textit{Proyectos de Generación de Conocimiento 2022} grant No. PID2022-137331OB-C33.


\bibliography{sn-bibliography}

\newpage
\begin{appendices}
    \section{Empirical assessment of convergence}

\begin{figure}[htb!]
    \centering
    \includegraphics[width=0.8\textwidth]{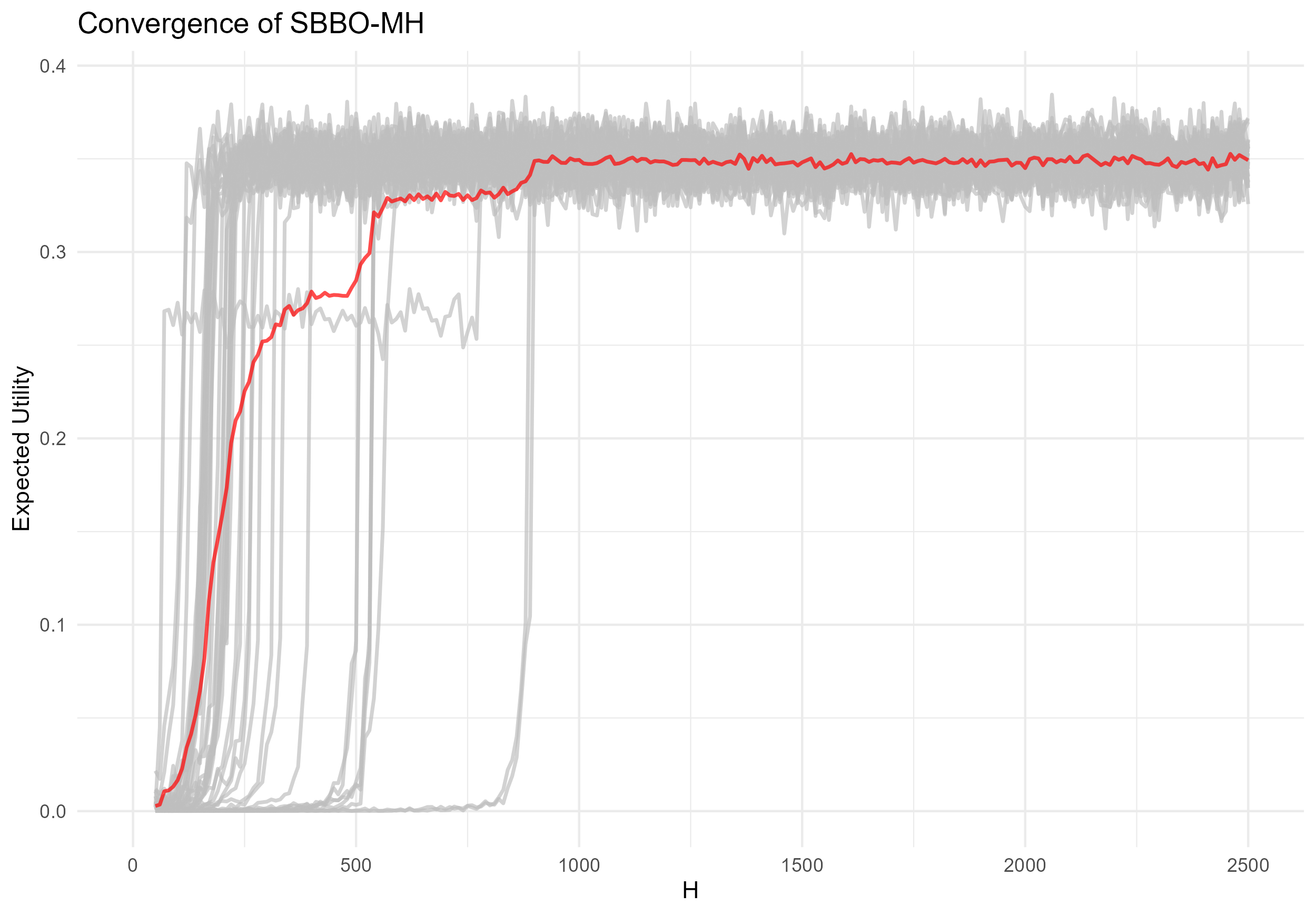}
    \caption{Posterior predictive utility versus $H$ for SBBO with Metropolis-Hastings.}
    \label{fig:convergence}
\end{figure}

    We include Figure \ref{fig:convergence} to suggest empirical guarantees of SBBO convergence. To make this plot, we initialized the contamination problem described in Section \ref{sec:cont_problem} with a dataset $\mathcal{D}_{1:100}$ of 100 of pairs of covariates and outcomes. Metropolis Hastings SBBO (Algorithm \ref{alg:sbbo}) with Tanimoto Gaussian Process regression as surrogate probabilistic model was used to find the subsequent evaluation location $x_{101}$. This process was repeated 40 times to account for the stochasticity of the SBBO algorithm. At each repetition, the cooling schedule incrementally increases $H$ from 50 to 2500 in steps of 10. For each $H$, we approximate the posterior predictive utility of the solution found by SBBO for that value of $H$ via Monte Carlo. Expected improvement was used as utility function. $H$ is represented in the $x$-axis and the corresponding posterior predictive utility in the $y$-axis. Each gray line corresponds to a repetition and the red line indicates the average.

\end{appendices}

\end{document}